\pdfoutput=1
\documentclass[conference,letterpaper,10pt]{IEEEtran}
\IEEEoverridecommandlockouts
\usepackage[T1]{fontenc}
\usepackage{cite}
\usepackage{amsmath,amssymb}
\usepackage{graphicx}
\usepackage{booktabs}
\usepackage{xcolor}
\usepackage[hidelinks]{hyperref}
\graphicspath{{figures/}}
\newcommand{\adMax}{35.3}
\newcommand{\adMin}{15.4}

\newcommand{\aucGainMax}{0.08}
\newcommand{\aucGainMin}{0.05}
\newcommand{\aucRelMax}{29}
\newcommand{\aucRelMin}{16}

\newcommand{\auconeThirt}{0.72}

\newcommand{\aucsixteenThirt}{0.80}

\newcommand{\callPmMax}{9.4}
\newcommand{\callPmMin}{2.8}
\newcommand{\callPrMax}{2.5}
\newcommand{\callPrMin}{2.2}
\newcommand{\candRecMax}{91.8}
\newcommand{\candRecMin}{87.0}
\newcommand{\cfNineFiveBestThirt}{15.8}
\newcommand{\cfNineFiveThirt}{33.6}
\newcommand{\cfTripBestThirt}{7.7}
\newcommand{\cfTripThirt}{28.2}
\newcommand{\cfbestFourt}{17.5}

\newcommand{\cfbestThirt}{11.3}

\newcommand{\cfoneThirt}{0.9}
\newcommand{\cfprofFourt}{32.0}

\newcommand{\cfprofThirt}{31.6}

\newcommand{\clHaiD}{22.0}
\newcommand{\clHaiDown}{0}
\newcommand{\clHaiFone}{0.0}
\newcommand{\clHaiFsix}{22.0}

\newcommand{\clHaiP}{1\times10^{-13}}
\newcommand{\clHaiRone}{0}
\newcommand{\clHaiRsix}{53}
\newcommand{\clHaiSnC}{7.5}
\newcommand{\clHaiSnP}{0.039}
\newcommand{\clHaiSnX}{14.2}
\newcommand{\clHaiUp}{44}

\newcommand{\clSonD}{24.0}
\newcommand{\clSonDown}{0}
\newcommand{\clSonFone}{0.5}
\newcommand{\clSonFsix}{24.5}

\newcommand{\clSonP}{7\times10^{-15}}

\newcommand{\clSonRFone}{4.5}
\newcommand{\clSonRFsix}{25.0}

\newcommand{\clSonRSnC}{13.2}
\newcommand{\clSonRSnP}{0.012}
\newcommand{\clSonRSnX}{24.5}

\newcommand{\clSonRone}{11}
\newcommand{\clSonRsix}{71}
\newcommand{\clSonSnC}{11.3}
\newcommand{\clSonSnP}{0.146}
\newcommand{\clSonSnX}{17.0}
\newcommand{\clSonUp}{48}

\newcommand{\clfourThirt}{30.8}

\newcommand{\clsixteenThirt}{41.3}

\newcommand{\coDownfourThirt}{5}

\newcommand{\coPmax}{2\times10^{-5}}

\newcommand{\coUpfourThirt}{72}

\newcommand{\cofourThirt}{55.1}

\newcommand{\cosixteenThirt}{56.2}

\newcommand{\cvAttMis}{97}

\newcommand{\cvRoneFourt}{22.2}
\newcommand{\cvRoneMis}{2.6}
\newcommand{\cvRoneOlmo}{14.3}
\newcommand{\cvRoneQthree}{42.5}
\newcommand{\cvRoneSev}{22.6}
\newcommand{\cvRsixFourt}{23.9}
\newcommand{\cvRsixMis}{4.5}
\newcommand{\cvRsixOlmo}{10.0}
\newcommand{\cvRsixQthree}{51.6}
\newcommand{\cvRsixSev}{15.4}

\newcommand{\dfmrMax}{32.3}
\newcommand{\dfmrMin}{13.5}

\newcommand{\disoneFourt}{2.5}

\newcommand{\disoneThirt}{2.1}
\newcommand{\dissixteenFourt}{24.1}

\newcommand{\dissixteenThirt}{23.4}

\newcommand{\eFpmaxQthree}{0.55}

\newcommand{\eFprofQthree}{0.53}

\newcommand{\ePancFourt}{0.74}

\newcommand{\ePancThirt}{0.76}

\newcommand{\ePprofFourt}{0.61}

\newcommand{\ePprofThirt}{0.75}

\newcommand{\eRancThirt}{0.14}
\newcommand{\eRatioMax}{2.7}
\newcommand{\eRatioMin}{1.4}

\newcommand{\eRprofMis}{0.02}

\newcommand{\eRprofThirt}{0.32}

\newcommand{\entDownThirt}{1}
\newcommand{\entMaxAgents}{313}
\newcommand{\entMedAgents}{28}
\newcommand{\entTieThirt}{139}

\newcommand{\entUpThirt}{126}
\newcommand{\exSameLoc}{4}
\newcommand{\excMaxFourt}{17.3}
\newcommand{\excMaxMis}{16.7}
\newcommand{\excMaxOlmo}{2.4}
\newcommand{\excMaxQthree}{6.4}

\newcommand{\excMaxThirt}{15.2}
\newcommand{\excMinFourt}{11.3}
\newcommand{\excMinMis}{2.8}

\newcommand{\excMinThirt}{6.0}

\newcommand{\fmroneMis}{0.4}

\newcommand{\fmroneThirt}{4.5}
\newcommand{\fmrsixteenFourt}{37.6}
\newcommand{\fmrsixteenMis}{24.1}

\newcommand{\fmrsixteenThirt}{36.8}

\newcommand{\fxAudProf}{\textminus{}0.06}
\newcommand{\fxCalN}{9}

\newcommand{\fxCallsAud}{3.00}
\newcommand{\fxCallsProf}{2.37}
\newcommand{\fxDPmax}{+1.2}
\newcommand{\fxDPmin}{\textminus{}4.7}
\newcommand{\fxMisRone}{1.6}
\newcommand{\fxMisRtwo}{2.9}
\newcommand{\fxNiMax}{10.3}
\newcommand{\fxNiMin}{1.2}

\newcommand{\fxQthreeRatio}{0.98}
\newcommand{\fxRRmax}{2.1}
\newcommand{\fxRRmin}{1.3}
\newcommand{\fxRsig}{5}
\newcommand{\fxSoftDP}{0.4}

\newcommand{\hardF}{0.76}
\newcommand{\hardT}{0.60}

\newcommand{\holmN}{56}

\newcommand{\hsFn}{331}
\newcommand{\hsFoneMax}{19.3}
\newcommand{\hsFoneMin}{0.0}
\newcommand{\hsFsixMax}{45.6}
\newcommand{\hsFsixMin}{22.1}
\newcommand{\hsTn}{60}
\newcommand{\hsToneMax}{56.7}
\newcommand{\hsToneMin}{41.7}

\newcommand{\hsTsixMax}{51.7}
\newcommand{\hsTsixMin}{43.3}

\newcommand{\lastfThirt}{29.9}
\newcommand{\lnDown}{5}
\newcommand{\lnMisOne}{0.4}
\newcommand{\lnMisRep}{1.5}
\newcommand{\lnSetMax}{37.8}
\newcommand{\lnSetMin}{17.7}
\newcommand{\lnThirtOne}{4.5}
\newcommand{\lnThirtRep}{1.7}

\newcommand{\loAgree}{99.9}

\newcommand{\loLoMax}{25}
\newcommand{\loLoMin}{17}
\newcommand{\loRoneThirt}{18.0}
\newcommand{\loRsixThirt}{10.7}

\newcommand{\massMin}{0.902}

\newcommand{\mcDownThirt}{1}

\newcommand{\mcPThirt}{1\times10^{-50}}
\newcommand{\mcPmax}{5\times10^{-19}}

\newcommand{\mcUpThirt}{173}

\newcommand{\mitBoundMax}{8}
\newcommand{\mitSkepMax}{12}
\newcommand{\mitSkepMaxModel}{Qwen2.5-14B}

\newcommand{\nBlockRecords}{2{,}572}
\newcommand{\nBlockSirens}{404}
\newcommand{\nBlocks}{80}
\newcommand{\nEnt}{266}

\newcommand{\nExactA}{61}
\newcommand{\nExactB}{36}

\newcommand{\nNeg}{532}
\newcommand{\nPos}{266}
\newcommand{\nPrompts}{16{,}758}
\newcommand{\nSecond}{276}
\newcommand{\nSiren}{162{,}104}
\newcommand{\nSiret}{203{,}988}
\newcommand{\nSupplier}{275{,}865}
\newcommand{\nTests}{18}
\newcommand{\negABpMax}{8\times10^{-10}}
\newcommand{\negAoneMax}{16.2}
\newcommand{\negAoneMin}{0.0}
\newcommand{\negAsixMax}{42.1}
\newcommand{\negAsixMin}{19.2}
\newcommand{\negBsixMax}{33.1}
\newcommand{\negBsixMin}{16.2}
\newcommand{\ordDfmrMax}{+0.8}
\newcommand{\ordDfmrMin}{\textminus{}0.8}

\newcommand{\ornsixteenFourt}{25.4}
\newcommand{\ornsixteenMis}{7.3}

\newcommand{\ornsixteenThirt}{23.1}

\newcommand{\pEntBonf}{0.002}

\newcommand{\pEntMax}{1\times10^{-4}}

\newcommand{\pEntThirt}{2\times10^{-36}}
\newcommand{\pSiret}{73.9}
\newcommand{\popAgents}{5.1}
\newcommand{\popLots}{9.7}
\newcommand{\popSingle}{88}

\newcommand{\pvSig}{16}

\newcommand{\recOneMaxAll}{42}
\newcommand{\recOneMinAll}{3}

\newcommand{\reconeThirt}{18.4}

\newcommand{\recsixteenThirt}{87.6}

\newcommand{\satNegThirt}{12.4}
\newcommand{\satOnFourt}{85}
\newcommand{\satOnThirt}{90}
\newcommand{\satOnThirtWrong}{83}

\newcommand{\satPosThirt}{46.2}

\newcommand{\sgOtherMaxAll}{16.1}

\newcommand{\sgOtherMinAll}{0.9}

\newcommand{\sysLlmcerMax}{+0.1}
\newcommand{\sysLlmcerMin}{\textminus{}39.5}

\newcommand{\tedAucMax}{0.93}
\newcommand{\tedAucMin}{0.91}

\newcommand{\tedPool}{17{,}267}
\newcommand{\tedRecMax}{99}
\newcommand{\tedRecMin}{77}
\newcommand{\tedSevD}{\textminus{}9.3}
\newcommand{\tedSnN}{12}
\newcommand{\tedSnSig}{11}
\newcommand{\tedSnThirtC}{2.4}
\newcommand{\tedSnThirtX}{67.5}

\begin{document}

\title{Growing Clusters Lower the Bar:\\Profile Accretion in LLM-Based\\Incremental Entity Resolution}

\newif\ifanon \anonfalse
\ifanon
\author{\IEEEauthorblockN{Anonymous Author(s)}
\IEEEauthorblockA{Affiliation withheld for review}}
\else
\author{\IEEEauthorblockN{Huixue Han\textsuperscript{1},
Shangzhi Xu\textsuperscript{2},
Anqi Xiong\textsuperscript{1} and
Yu-Chu Tian\textsuperscript{1,*}}
\IEEEauthorblockA{\textsuperscript{1}\textit{School of Computer Science, Queensland University of Technology, Brisbane, Australia}}
\IEEEauthorblockA{\textsuperscript{2}\textit{Independent Researcher, Brisbane, Australia}}
\IEEEauthorblockA{n12101419@qut.edu.au, mxu490469@gmail.com, anqi.xiong@hdr.qut.edu.au, y.tian@qut.edu.au\\
\textsuperscript{*}Corresponding author}}
\fi

\maketitle

\begin{abstract}
Incremental entity resolution pipelines ask a large language model (LLM) whether a new record belongs to an existing cluster and show it the records already linked. Analyses of matching oracles assume that the answer depends only on the records compared. For the LLM matchers we test, it depends on cluster size, and the model lowers its bar for joining a cluster as the cluster grows. On French procurement suppliers from FOPPA, with SIREN identifiers as ground truth, we hold the new record fixed and vary only the number of verified members shown. Sixteen members instead of one raise the false match rate on similarly named companies by \dfmrMin--\dfmrMax{} percentage points in six open-weight models, while their area under the ROC curve rises by only \aucGainMin--\aucGainMax. In Claude Sonnet~5 and Haiku~5.5 the rise is \clSonD{} and \clHaiD{} points. The shift persists with street addresses, prompt variants and the BatchER and ComEM formats, and one false merge raises acceptance of the intruder's other records in every open model. It does not arise on Czech and Bulgarian TED notices, where one record with its full address already identifies the company. In online linking on confusable blocks, a single threshold set on labelled data raises recall by a factor of \fxRRmin--\fxRRmax{} for five models with no detectable loss of precision and no extra LLM calls.
\end{abstract}


\section{Introduction}

Large language models (LLMs) increasingly judge new inputs against a reference set that grows as they work, as judges comparing against earlier verdicts~\cite{zheng2023judging} or as matchers comparing a record with the records already linked. Whether the size of that set changes the judgment, with the input held fixed, has not been measured. We study the question in entity resolution, where the reference set is a cluster and ground truth can come from registry identifiers.

Entity resolution (ER) decides which records refer to the same real-world entity~\cite{fellegi1969theory,elmagarmid2007duplicate,getoor2012entity,christen2012data,christophides2021overview}. In public procurement it determines which supplier won which contracts. Concentration statistics, indicators of corruption risk and procurement knowledge graphs depend on it~\cite{fazekas2016objective,fazekas2020uncovering,soylu2022theybuyforyou,potin2023foppa}, and because supplier names vary and identifiers are often missing, resolving suppliers depends on matching names and addresses.

Matchers have moved from feature-based systems~\cite{konda2016magellan} to deep and pre-trained language models~\cite{mudgal2018deep,li2020ditto}, and large language models now match records well without task-specific training~\cite{narayan2022wrangle,peeters2025entity} or after modest fine-tuning~\cite{steiner2025finetuning,zhang2024jellyfish}. Attention has shifted to how judgments are combined into clusters~\cite{fan2024batcher,wang2025match,fu2025llmcer}. In incremental designs each new record is compared with existing clusters, and a cluster is represented to the model by the records already linked to it, a representation that grows with every link~\cite{benjelloun2009swoosh,gruenheid2014incremental,whang2014incremental}. The analyses that tell such pipelines which comparisons to make, and what accuracy to expect, treat the matcher as an oracle whose answer depends only on the records compared~\cite{wang2013transitive,firmani2016online,galhotra2018robust}. An LLM that is shown a cluster profile need not behave this way.

We measure this effect, which we call \emph{profile accretion}, and find that the matcher's criterion moves with the cluster: the larger the cluster, the less evidence a new record needs to join it. Measuring this needs a query held fixed, profiles that grow in nested steps and ground truth that does not rely on annotation, which benchmarks of fixed pairs or sets do not provide. We use supplier records from FOPPA~\cite{potin2023foppa}, in which the SIREN number identifies the legal entity. For \nEnt{} legal entities we show the model one to sixteen verified records of the entity together with a held-out record of the same entity or a record of a different entity with a similar name. We repeat the design with two closed models, with an added address field, with three prompt variants, in the matching formats of three published LLM-ER systems, and on a second dataset from TED keyed by registry identifiers. Our contributions are:
\begin{itemize}
\item \textbf{Accretion shifts the decision criterion.} From one to sixteen members, the false match rate rises by \dfmrMin--\dfmrMax{} points for six open models (three families) and by \clSonD{} and \clHaiD{} points for two Claude models, while AUC rises by \aucGainMin--\aucGainMax{} (Sections~\ref{sec:accretion}, \ref{sec:scope}).
\item \textbf{Profiles create matches no member supports.} Three models accept more than the OR of their single member judgments, two with confident errors (Section~\ref{sec:conf}).
\item \textbf{A causal snowball.} At equal profile size, one falsely merged record raises acceptance of the intruder's other records for every open model on both datasets (Sections~\ref{sec:snowball}, \ref{sec:scope}).
\item \textbf{A predictable scope.} The shift holds with addresses, under prompt variants and in the BatchER and ComEM formats. It arises where single records are weak evidence of identity and is absent where one record already identifies the entity; in-context clustering produces the opposite error (Section~\ref{sec:scope}).
\item \textbf{A simple remedy.} Aggregate $F_1$ rewards the shift. In a stress test, one threshold set on labelled data for the required precision raises recall by \fxRRmin--\fxRRmax{} times for five models with no detectable loss of precision and no extra LLM calls. Calibration per cluster size and cluster audits add nothing to it (Sections~\ref{sec:e2e}, \ref{sec:fix}).
\end{itemize}

\section{Related Work}

\textbf{LLM-based matching.} Batching and demonstration selection reduce the cost of LLM matching~\cite{fan2024batcher}, the order of demonstrations changes the matching accuracy of small models~\cite{arvanitis2025entity}, showing several candidates at once improves accuracy~\cite{wang2025match}, and in-context clustering asks the model to group sets of records with guardrails against misclustering~\cite{fu2025llmcer}. Evaluations, from early benchmarks~\cite{kopcke2010evaluation} to LLM studies, score fixed pairs or sets. None holds the query fixed while varying, in nested steps, the members shown for its cluster.

\textbf{LLM confidence and context.} Neural networks are often miscalibrated, pre-trained transformers less so in domain, and language-model confidence is informative but imperfect~\cite{guo2017calibration,desai2020calibration,jiang2021calibration}. In-context predictions shift with the label balance and order of the examples in a prompt~\cite{zhao2021calibrate,lu2022fantastically}, irrelevant context degrades reasoning~\cite{shi2023distracted}, the use of long contexts depends on the position of information~\cite{liu2024lost}, and LLM judges show position and verbosity biases~\cite{zheng2023judging} and drift toward prior scores shown as context~\cite{kapetanovic2026anchoring}. A cluster profile is context of this kind whose size and content change with every link.

\textbf{Clustering in ER.} Probabilistic record linkage scores record pairs~\cite{fellegi1969theory}, blocking restricts which pairs are scored~\cite{papadakis2020blocking}, and a clustering step turns pairwise decisions into entities~\cite{christophides2021overview}. Transitive closure lets one false edge join unrelated clusters, while center-based clustering compares records only with a representative and under-merges~\cite{hassanzadeh2009framework,saeedi2017comparative}. Collective ER uses decisions on related references as evidence for further decisions~\cite{bhattacharya2007collective}. The Swoosh family merges matching records and re-compares the merged record with all others, which can create matches that no single source record supports~\cite{benjelloun2009swoosh}. Incremental ER updates clusters as records are inserted and can split or repair them~\cite{gruenheid2014incremental,whang2014incremental}. Clusterings are evaluated with pairwise and cluster-level measures~\cite{menestrina2010evaluating}.

\textbf{Matchers as oracles.} A line of work treats the pairwise matcher, whether a crowd or a model, as an oracle and asks which pairs to query and how to exploit transitivity~\cite{wang2012crowder,wang2013transitive,vesdapunt2014crowdsourcing,firmani2016online}; robust variants tolerate noisy answers~\cite{galhotra2018robust}. In these analyses the answer to a query depends on the records compared, not on what a cluster already contains. Profile accretion violates this assumption.

\begin{figure*}[t]
\centering
\includegraphics[width=\textwidth]{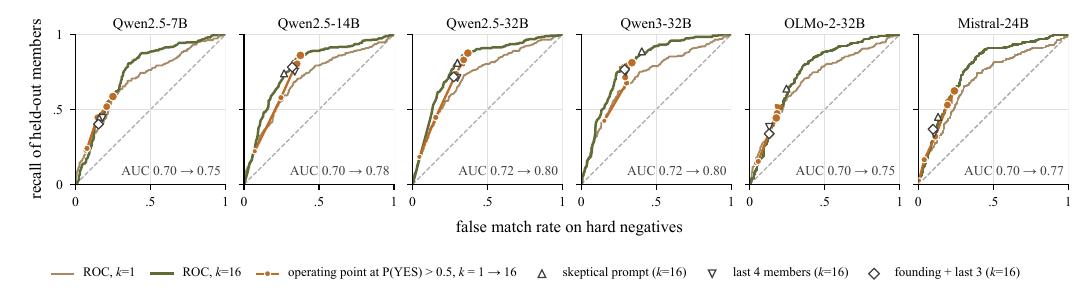}
\caption{As the profile grows, the operating point moves far along a nearly unchanged ROC curve (\nEnt{} FOPPA legal entities, one panel per model). Curves: ROC for separating the \nPos{} held-out true members from the \nNeg{} hard negatives (A and B pooled) with one verified member ($k=1$) and with 16 ($k=16$). Connected points: the operating point at $P(\text{YES})>0.5$ for $k=1,2,4,8,16$; open markers: the three mitigations at $k=16$.}
\label{fig:accretion}
\end{figure*}

\begin{table*}[t]
\centering
\caption{Controlled accretion results (\%, except AUC). FMR: false match rate on \nNeg{} hard negatives; OR null: disjunction of single member judgments at $k=16$; distinct names: negatives whose name differs from every member ($k=1$\,/\,$k=16$); Conf.: share of hard negatives with $P(\text{YES})>0.99$ for the most similar member judged alone and for the profile (Fig.~\ref{fig:mitig}b); matched recall: recall at $k=1$ and at $k=16$ with the $k=16$ threshold cross-validated over entities so that the held-out FMR stays at or below its $k=1$ value ($^\dagger$: log-odds scale, as $P(\text{YES})$ rounds to 1). Differences in the text are computed before rounding.}
\label{tab:main}
\setlength{\tabcolsep}{4.2pt}
\begin{tabular}{lccccccccccc}
\toprule
 & \multicolumn{4}{c}{False matches} & \multicolumn{2}{c}{Recall} & \multicolumn{2}{c}{AUC} & \multicolumn{2}{c}{Conf. ($k=16$)} & Matched recall \\
\cmidrule(lr){2-5}\cmidrule(lr){6-7}\cmidrule(lr){8-9}\cmidrule(lr){10-11}
Model & $k=1$ & $k=16$ & OR null & distinct names & $k=1$ & $k=16$ & $k=1$ & $k=16$ & best member & profile & ($k$=1\,/\,16) \\
\midrule
Qwen2.5-7B & 6.8 & 25.0 & 29.3 & 2.8\,/\,13.1 & 22.6 & 58.6 & 0.70 & 0.75 & 14.1 & 12.4 & 22.6\,/\,15.4 \\
Qwen2.5-14B & 7.1 & 37.6 & 25.4 & 2.5\,/\,24.1 & 22.2 & 86.1 & 0.70 & 0.78 & 17.5 & 32.0 & 22.2\,/\,23.9 \\
Qwen2.5-32B & 4.5 & 36.8 & 23.1 & 2.1\,/\,23.4 & 18.4 & 87.6 & 0.72 & 0.80 & 11.3 & 31.6 & 18.0\,/\,10.7$^\dagger$ \\
Qwen3-32B & 15.4 & 33.8 & 40.2 & 8.5\,/\,20.0 & 42.5 & 81.2 & 0.72 & 0.80 & 23.3 & 21.8 & 42.5\,/\,51.6 \\
OLMo-2-32B & 4.1 & 17.7 & 21.1 & 0.7\,/\,8.3 & 14.3 & 44.4 & 0.70 & 0.75 & 1.5 & 0.9 & 14.3\,/\,10.0 \\
Mistral-24B & 0.4 & 24.1 & 7.3 & 0.2\,/\,11.3 & 2.6 & 62.4 & 0.70 & 0.77 & 0.0 & 0.0 & 2.6\,/\,4.5 \\

\bottomrule
\end{tabular}
\end{table*}

\section{Data and Design}

\subsection{Ground Truth from FOPPA}
FOPPA v1.1.3~\cite{potin2023foppa} lists the agents that appear in French award notices published in the EU's Tenders Electronic Daily from 2010 to 2020. Of the \nSupplier{} agents that won at least one lot, \nSiret{} (\pSiret\%) carry a valid 14-digit SIRET establishment number; its first nine digits form the SIREN, which identifies the legal entity. These agents belong to \nSiren{} legal entities, \popSingle\% of which have a single agent. Establishments of one company share a SIREN, while subsidiaries in a group have their own. A record is shown as one of the agent's raw name variants, drawn once per agent and fixed across all conditions and models, followed by its city and postcode. The identifiers are never shown.

\subsection{Controlled Profiles}
We use every legal entity with at least 17 agents (\nEnt{} entities; median \entMedAgents{}, maximum \entMaxAgents{} agents). They hold \popAgents\% of the supplier agents with a valid SIRET and \popLots\% of the lots these agents won. For each entity we order its agents at random and take the first 16 as verified members and the 17th as a held-out positive. Two hard negatives come from other entities: \emph{negative A} is most similar in name to the first (founding) member, \emph{negative B} to one of members 9--16; similarity is the cosine of character 3--4-gram TF-IDF vectors over the agents' cleaned canonical names~\cite{gravano2003text}. In \nExactA{} (negative A) and \nExactB{} (negative B) cases the negative's displayed name equals a member's name. Only \exSameLoc\% of these also share that member's city and postcode, as a SIRET coding error would. We report negatives whose name differs from every member separately as \emph{distinct-name} negatives.

For each query we build profiles with the first $k\in\{1,2,4,8,16\}$ members, so larger profiles contain the smaller ones. The founding member is always listed first. As a control for position, the $k=16$ profile is also shown in reverse order.

\subsection{Matching Protocol}
The system prompt defines the target: records refer to the same legal entity if they share a SIREN; branches of one company count as the same, and distinct legal entities of a group count as different. The user turn lists the profile as ``records verified to belong to one legal entity'', gives the new record and asks for a single word, YES or NO. We read the next-token distribution after the chat template and define $P(\text{YES})$ as the probability mass on YES tokens (any case) divided by the mass on YES and NO tokens. This mass was at least \massMin{} for every prompt of the main design, so the models answered within the two options. A match is declared at $P(\text{YES})>0.5$.

We evaluate Qwen2.5-7B-, 14B- and 32B-Instruct~\cite{qwen25report} and Qwen3-32B~\cite{qwen3report}. Qwen3 runs with thinking disabled, so that its first token is the answer. Four checkpoints thus share the Qwen lineage. Two further families were trained independently of Qwen: OLMo-2-0325-32B-Instruct~\cite{olmo2025furious}, a fully open model with published training data and recipe, and Mistral-Small-24B-Instruct-2501 from Mistral AI. Models run in bfloat16 on one H100 GPU (OLMo-2 sharded over two A100 GPUs), over \nPrompts{} prompts per model. Every query is also judged against each of the 16 members alone. The disjunction of these single member judgments over the first $k$ members, the \emph{OR null}, is the decision of a matcher that links whenever any one member is convincing.

\subsection{Analysis Framework}
We read the results through signal detection theory~\cite{macmillan2004detection}. A matcher shown a profile of size $k$ separates held-out members from hard negatives with some sensitivity and accepts above some criterion. The area under the ROC curve (AUC)~\cite{fawcett2006roc} measures sensitivity independently of the threshold, whereas the false match rate and recall at a fixed threshold depend on both. A joint rise of both rates with little change in AUC therefore indicates a shift of the criterion: the operating point moves along a nearly unchanged ROC curve (Fig.~\ref{fig:accretion}). Tversky's contrast model of similarity~\cite{tversky1977features}, in which similarity increases with shared features and decreases with distinctive ones, predicts such a shift whenever the judgment weighs features shared with any member more heavily than profile features that the new record lacks. Each added member contributes features a negative may share, such as a legal form, a city or a word of the name, while the negative's own distinctive features stay fixed. Paired differences between conditions on the same queries are tested with the exact McNemar test~\cite{mcnemar1947note}. Because the two negatives of an entity share its members, we also test the per-entity change with an exact sign test and estimate 95\% intervals with a bootstrap over entities (2{,}000 resamples). For matched recall, the threshold at $k=16$ is chosen by five-fold cross-validation over entities (20 repetitions): it is set on the training entities so that their false match rate stays at or below their $k=1$ rate, and recall is measured on held-out entities.

\subsection{Interventions}
To test mitigations at $k=16$ we (i) add an instruction to check for evidence of a different legal entity before answering (\emph{skeptical prompt}), (ii) show only the last four members, and (iii) show the founding member and the last three. To test contamination we take, for each negative whose entity has another agent, a second record of that intruder entity (\nSecond{} queries). We compare a clean profile of $k$ verified members with a contaminated profile of $k-1$ verified members plus the intruder, for $k\in\{4,16\}$, so profile size is identical. As a length control, the founding record is repeated $k\in\{4,16\}$ times, which lengthens the profile without adding content.

\subsection{Online Linking}
For an end-to-end check we form \nBlocks{} blocks of confusable records. Each block contains up to 24 agents of one of the first 80 entities of the random pilot draw. It adds up to 8 agents of each of up to five other entities whose names have a cosine of at least 0.5 with that entity (\nBlockRecords{} records of \nBlockSirens{} legal entities in total). Records arrive in random order. Each is compared with the three existing clusters whose members are most similar by character trigrams (cosine $\geq0.3$), joins the cluster with the highest score if $P(\text{YES})>0.5$ and opens a new one otherwise. A cluster is represented by its founding member only, by its whole profile (founding member plus the 15 most recent members), or by the maximum of single member judgments (\emph{pairwise max}). We report pairwise precision, recall and $F_1$ against SIREN~\cite{menestrina2010evaluating}.

\subsection{Replications}\label{sec:repl}
\emph{Closed models.} Claude Sonnet~5 (\texttt{claude-sonnet-5}) and Claude Haiku~5.5 (\texttt{claude-haiku-5-5}), both with extended thinking disabled, receive the same prompts for a random 100 of the entities at $k\in\{1,4,16\}$, together with the snowball pairs at $k=4$. Token probabilities are not available, so each prompt is sampled three times at the API default temperature (no temperature set). A match is a majority of YES among the valid samples (ties count as NO), and a YES in all three samples counts as confident.
\emph{Fields and wording.} We rerun the design with each agent's street address added to its record. We also run three prompt variants at $k\in\{1,4,16\}$: a reworded instruction (V1), members shown as a numbered table with an ``Answer:'' readout (V2), and the bare question without the SIREN definition (V3).
\emph{Published formats.} We reimplement the matching formats of three LLM-based ER systems on the same queries. BatchER~\cite{fan2024batcher} asks batched pairwise questions with fixed demonstrations, which we aggregate by OR over members. ComEM~\cite{wang2025match} selects the query's cluster among two candidate clusters and a ``none'' option. LLM-CER~\cite{fu2025llmcer} clusters the members, the query and two distractor records in context. We reproduce each system's prompting format, the component that determines what the model sees of a cluster.
\emph{Second dataset.} From TED contract-award notices of 2014--2017 we take winners from Czechia and Bulgaria, whose national company identifiers (I\v{C}O, EIK) are filled in for about a third of winner rows and pass the national checksum in over 98\% of cases. A record shows name, street address, postcode, town and country. The 102 identifiers with at least 17 distinct records (16 Czech, 86 Bulgarian) enter the same design, with hard negatives drawn from all \tedPool{} distinct checksum-valid records. Details are in the online supplement.

\section{Results}

\begin{figure*}[t]
\centering
\includegraphics[width=\textwidth]{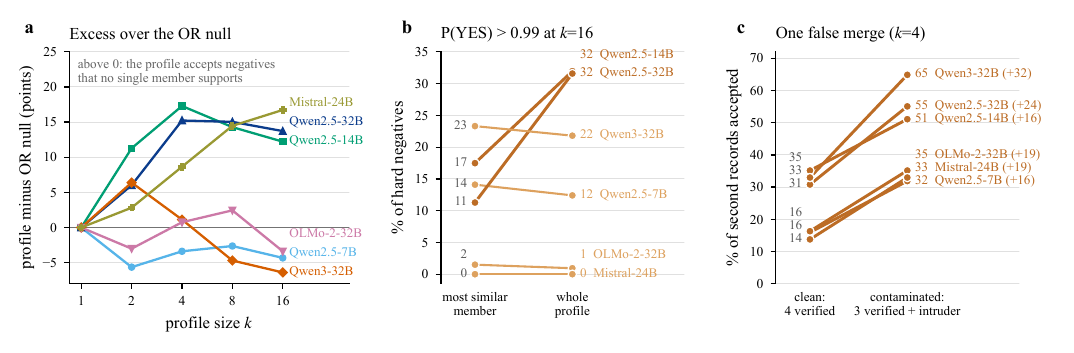}
\caption{Three models accept negatives that no member supports, two of them with confidence, and one false merge raises acceptance of the intruder's records in every model. (a) False match rate of the profile minus that of the OR null, the disjunction of the model's own single member judgments over the same members; above 0, the profile accepts negatives that no shown member supports alone. (b) Share of hard negatives with $P(\text{YES})>0.99$ for the most similar of the 16 members judged alone and for the 16-member profile. (c) Share of second records of an intruder's entity judged to be the same entity when the profile holds 4 verified members (clean) or 3 verified members plus the intruder (contaminated; \nSecond{} paired queries).}
\label{fig:mitig}
\end{figure*}

\begin{table*}[t]
\centering
\caption{Snowball and online linking. Left: share (\%) of second records of an intruder's entity judged to be the same entity, with a clean profile of $k$ verified members or with $k-1$ verified members plus the intruder. Right: pairwise precision / recall / $F_1$ of online linking on \nBlocks{} confusable blocks (\nBlockRecords{} records) for three cluster representations.}
\label{tab:e2e}
\setlength{\tabcolsep}{4pt}
\begin{tabular}{lcccc|ccc}
\toprule
 & \multicolumn{2}{c}{$k=4$} & \multicolumn{2}{c|}{$k=16$} & \multicolumn{3}{c}{Online linking (P / R / $F_1$)} \\
\cmidrule(lr){2-3}\cmidrule(lr){4-5}\cmidrule(lr){6-8}
Model & clean & contaminated & clean & contaminated & founding member & pairwise max & whole profile \\
\midrule
Qwen2.5-7B & 15.9 & 31.9 & 25.7 & 34.8 & 0.72 / 0.20 / 0.32 & 0.67 / 0.29 / 0.41 & 0.78 / 0.29 / 0.42 \\
Qwen2.5-14B & 35.1 & 51.1 & 38.8 & 52.5 & 0.74 / 0.17 / 0.27 & 0.71 / 0.26 / 0.38 & 0.61 / 0.45 / 0.52 \\
Qwen2.5-32B & 30.8 & 55.1 & 41.3 & 56.2 & 0.76 / 0.14 / 0.23 & 0.76 / 0.22 / 0.34 & 0.75 / 0.32 / 0.45 \\
Qwen3-32B & 33.0 & 64.9 & 37.3 & 60.5 & 0.70 / 0.30 / 0.42 & 0.65 / 0.47 / 0.55 & 0.66 / 0.44 / 0.53 \\
OLMo-2-32B & 16.3 & 35.1 & 20.7 & 35.5 & 0.77 / 0.15 / 0.25 & 0.78 / 0.18 / 0.29 & 0.77 / 0.23 / 0.36 \\
Mistral-24B & 13.8 & 33.0 & 27.2 & 41.7 & 0.89 / 0.02 / 0.03 & 0.90 / 0.02 / 0.04 & 0.90 / 0.02 / 0.05 \\

\bottomrule
\end{tabular}
\end{table*}

\subsection{Accretion Shifts the Decision Criterion}\label{sec:accretion}
Showing more verified members makes every model more willing to say YES (Fig.~\ref{fig:accretion}; Table~\ref{tab:main}). From $k=1$ to $k=16$ the false match rate rises for every model, for example from \fmroneThirt\% to \fmrsixteenThirt\% for Qwen2.5-32B and from \fmroneMis\% to \fmrsixteenMis\% for Mistral-24B, the most conservative model with a single member. In paired tests on the same negatives, \mcUpThirt{} judgments change from NO to YES and \mcDownThirt{} from YES to NO for Qwen2.5-32B ($p=\mcPThirt$, exact McNemar test~\cite{mcnemar1947note}); the weakest contrast for any model still has $p=\mcPmax$. At the entity level the false match rate rises for \entUpThirt{} entities, falls for \entDownThirt{} and is unchanged for \entTieThirt{} ($p=\pEntThirt$, sign test). Across models the increase is \dfmrMin--\dfmrMax{} points, and its 95\% bootstrap interval excludes zero for every model (Table~S1 of the online supplement). The \nTests{} sign tests over entities for accretion and contamination remain significant after Bonferroni correction (largest adjusted $p=\pEntBonf$). A Holm correction over all \holmN{} McNemar tests per model behind Table~\ref{tab:checks} leaves every count of significant increases unchanged.

The effect is not confined to near-duplicates: among distinct-name negatives the false match rate grows from \disoneThirt\% to \dissixteenThirt\% for Qwen2.5-32B and from \disoneFourt\% to \dissixteenFourt\% for Qwen2.5-14B. Nor is it confined to negative B, whose resembled member enters only at $k=16$. Negative A resembles the founding member, which every profile contains, and its false match rate rises from \negAoneMin--\negAoneMax\% to \negAsixMin--\negAsixMax\% (negative B: \negBsixMin--\negBsixMax\% at $k=16$; every McNemar $p\leq\negABpMax$, Table~S12).

Recall rises in parallel, from \reconeThirt\% to \recsixteenThirt\% for Qwen2.5-32B. The area under the ROC curve for separating held-out members from hard negatives grows from \auconeThirt{} to \aucsixteenThirt{} for the same model, and by \aucGainMin--\aucGainMax{} across models, which closes \aucRelMin--\aucRelMax\% of the gap to perfect separation. Discrimination therefore improves, but far less than the acceptance rates move, and most of the change is a shift in the decision criterion. We cross-validate the threshold at $k=16$ so that the false match rate on held-out entities matches its $k=1$ value. Recall is then \cvRsixSev\% for Qwen2.5-7B, \cvRsixFourt\% for Qwen2.5-14B, \cvRsixOlmo\% for OLMo-2-32B and \cvRsixMis\% for Mistral-24B, against \cvRoneSev\%, \cvRoneFourt\%, \cvRoneOlmo\% and \cvRoneMis\% with the founding member alone; for Mistral-24B such a threshold exists in \cvAttMis\% of folds. Only Qwen3-32B gains clearly (\cvRoneQthree\% to \cvRsixQthree\%). For Qwen2.5-32B, \satNegThirt\% of hard negatives and \satPosThirt\% of held-out members receive $P(\text{YES})=1$ at $k=16$, because the probability rounds to 1 once the log-odds of YES against NO are above about 17. Rescored with the log-odds computed from the logits (agreement with the original decisions \loAgree\%), these cases spread over log-odds of \loLoMin--\loLoMax, a threshold exists, and matched recall is \loRsixThirt\% against \loRoneThirt\% with one member (Table~S18).

Judged alone, the other 15 members give a false match rate close to the founding member's (\sgOtherMinAll--\sgOtherMaxAll\%, Table~S1), and reversing the 16-member profile changes it by \ordDfmrMin{} to \ordDfmrMax{} points. Length alone does not produce the shift. Repeating the founding record 16 times lengthens the prompt like 16 members but adds no content. It lowers the false match rate for \lnDown{} models (Qwen2.5-32B: \lnThirtOne\% to \lnThirtRep\%) and raises it only for Mistral-24B (\lnMisOne\% to \lnMisRep\%). Sixteen distinct members give \lnSetMin--\lnSetMax\% in the same run (Table~S15). The shift thus depends on what the added members contain, as Tversky's contrast model predicts.

\subsection{Matches beyond the OR Null}\label{sec:conf}
A matcher that links whenever any one member convinces it would lie on the zero line of Fig.~\ref{fig:mitig}a. Qwen2.5-7B stays below this line, and OLMo-2-32B and Qwen3-32B lie above it by at most \excMaxOlmo{} and \excMaxQthree{} points at intermediate sizes and fall below it at $k=16$. Qwen2.5-14B and 32B lie above the OR null at every $k\geq2$, by \excMinFourt--\excMaxFourt{} and \excMinThirt--\excMaxThirt{} points (\fmrsixteenFourt\% and \fmrsixteenThirt\% at $k=16$, against \ornsixteenFourt\% and \ornsixteenThirt\%), and Mistral-24B by \excMinMis--\excMaxMis{} points (\fmrsixteenMis\% against \ornsixteenMis\%). These three models accept negatives that none of the shown members would have convinced them of on its own.

The excess comes with confidence (Fig.~\ref{fig:mitig}b). With the founding member alone, \cfoneThirt\% of hard negatives receive $P(\text{YES})>0.99$ from Qwen2.5-32B. The most similar of the 16 members, judged alone, raises this to \cfbestThirt\%; the 16-member profile raises it to \cfprofThirt\%. Qwen2.5-14B shows the same pattern (\cfbestFourt\% and \cfprofFourt\%); for the other models the profile is no more confident than the best single member. Lying above the OR null and confident error are therefore separate properties. The pattern holds at cutoffs of 0.95 and 0.999 (Qwen2.5-32B: \cfNineFiveBestThirt\% and \cfNineFiveThirt\%; \cfTripBestThirt\% and \cfTripThirt\%).

\subsection{Mitigations}\label{sec:mitig}
Fig.~\ref{fig:accretion} places each mitigation at $k=16$ in ROC space, next to the path of the plain profile from $k=1$ to $k=16$. Four members already produce most of the accretion, so bounded profiles move a model along its curve, within \mitBoundMax{} recall points. For Qwen2.5-32B the last four members give a false match rate of \lastfThirt\%, against \fmrsixteenThirt\% for the full profile and \fmroneThirt\% for one member. The skeptical prompt's operating point lies up to \mitSkepMax{} points above the curve for \mitSkepMaxModel{} but raises the false match rate of Qwen3-32B and OLMo-2-32B (Table~S2). Neither intervention removes the shift.

\subsection{Contamination by a False Merge}\label{sec:snowball}
Replacing one verified member with a falsely merged record changes how the model treats the intruder's entity (Fig.~\ref{fig:mitig}c; Table~\ref{tab:e2e}). With four verified members, a second record of the intruder's entity is accepted in \clfourThirt\% of cases for Qwen2.5-32B; after the swap the rate is \cofourThirt\% (\coUpfourThirt{} paired changes to YES, \coDownfourThirt{} back), and the increase holds for every model (Table~\ref{tab:e2e}). The effect persists with 16 members (\clsixteenThirt\% to \cosixteenThirt\% for Qwen2.5-32B), and the contrast for every model has $p\leq\coPmax$ (sign test over entities $p\leq\pEntMax$). Because the profiles have the same size and differ in one record, the increase is caused by the contamination itself. One error thus makes the next error of the same kind more likely.

\subsection{Online Linking}\label{sec:e2e}
For \candRecMin--\candRecMax\% of arriving records with a co-referent cluster, that cluster is among the candidates under every representation, so the differences arise in the matching decision. At the default threshold the accumulated profile raises recall by \eRatioMin{} to \eRatioMax{} times over the founding member representation for five of six models, for example from \eRancThirt{} to \eRprofThirt{} for Qwen2.5-32B, and gives a higher $F_1$ for all six models. It is the best of the three representations for all except Qwen3-32B, for which pairwise max leads (\eFpmaxQthree{} against \eFprofQthree; Table~\ref{tab:e2e}); BCubed $F_1$ also ranks the profile above the founding member for all six models (Table~S14). The shift appears in precision, which holds for Qwen2.5-32B (\ePancThirt{} and \ePprofThirt) but falls from \ePancFourt{} to \ePprofFourt{} for Qwen2.5-14B, the model with the largest excess over the OR null in the controlled setting. An evaluation by aggregate $F_1$ therefore selects the accumulated profile without showing that its precision at the default threshold depends on cluster size, which Section~\ref{sec:fix} addresses. (Mistral-24B's clusters seldom grow; its recall is \eRprofMis{} even with the profile.)

\begin{table}[t]
\centering
\caption{Scope of the effect. Change in the false match rate on hard negatives from $k=1$ to $k=16$ (range over models) and number of models whose increase is significant (95\% bootstrap interval above zero). Rows indented under FOPPA change one factor of the main design; ``-style'' rows reproduce a published prompting format with our models, so rates are not comparable with the original papers. On TED one model decreases significantly. Per-model results are in the online supplement.}
\label{tab:checks}
\footnotesize
\setlength{\tabcolsep}{3pt}
\begin{tabular}{lccc}
\toprule
Setting & models & $\Delta$FMR (points) & sig.\ $\uparrow$ \\
\midrule
FOPPA, name + city (main) & 6 & $+$13.5 to $+$32.3 & 6/6 \\
\quad + street address & 6 & $+$15.4 to $+$35.3 & 6/6 \\
\quad reworded prompt (V1) & 6 & $+$0.4 to $+$25.7 & 4/6 \\
\quad table + `Answer:' (V2) & 6 & $+$9.6 to $+$23.1 & 6/6 \\
\quad no entity definition (V3) & 6 & $+$18.0 to $+$32.1 & 6/6 \\
\quad BatchER-style batching & 6 & $+$18.4 to $+$31.8 & 6/6 \\
\quad ComEM-style selection & 6 & $+$21.8 to $+$43.2 & 6/6 \\
\quad LLM-CER-style clustering & 6 & $-$39.5 to $+$0.1 & 0/6 \\
Claude Sonnet 5, Haiku 5.5 & 2 & $+$22.0 to $+$24.0 & 2/2 \\
TED CZ/BG, full address & 6 & $-$9.3 to $+$2.0 & 0/6 \\

\bottomrule
\end{tabular}
\end{table}

\subsection{Scope of the Effect}\label{sec:scope}
Table~\ref{tab:checks} summarizes the replications. The two closed models, both run with thinking disabled, move like the open ones. The false match rate of Claude Sonnet~5 rises from \clSonFone\% at $k=1$ to \clSonFsix\% at $k=16$ (\clSonUp{} changes from NO to YES and \clSonDown{} in the other direction, $p=\clSonP$). That of Haiku~5.5 rises from \clHaiFone\% to \clHaiFsix\% (\clHaiUp{} and \clHaiDown, $p=\clHaiP$). Recall rises from \clSonRone\% to \clSonRsix\% and from \clHaiRone\% to \clHaiRsix\%. With its default reasoning Sonnet~5 shifts as well (\clSonRFone\% to \clSonRFsix\%). One false merge raises acceptance of the intruder's second record from \clHaiSnC\% to \clHaiSnX\% for Haiku~5.5 ($p=\clHaiSnP$) and from \clSonRSnC\% to \clSonRSnX\% for Sonnet~5 with reasoning ($p=\clSonRSnP$), and without reasoning in the same direction (\clSonSnC\% to \clSonSnX\%, $p=\clSonSnP$).

Adding the street address to every record leaves the shift intact (\adMin--\adMax{} points, significant for all six models). Under the three prompt variants, \pvSig{} of 18 combinations of model and variant show a significant increase; the exceptions, OLMo-2-32B and Mistral-24B under V1, accept almost no record at any size (recall below 1\% at $k=16$). In the ComEM format the rise is a change in one choice among clusters. In the BatchER format the decision is the OR of the members' answers, and most of the rise comes from the added questions. The founding member's own answer moves toward YES for three models, toward NO for Mistral-24B and not for two (Table~S7). Both formats err more as the cluster grows.

The effect is absent in two settings. In-context clustering in the style of LLM-CER lowers both the false match rate and recall as the record set grows (\sysLlmcerMin{} to \sysLlmcerMax{} points). Asked to partition a larger set, the models leave the query outside the profile's group more often. This is under-merging, not accretion. On TED, the false match rate does not rise for any model (\tedSevD{} to $+$2.0 points) and falls significantly for Qwen2.5-7B. There one record with street address and postcode already identifies the company. Recall with the founding member alone is \tedRecMin--\tedRecMax\% (FOPPA: \recOneMinAll--\recOneMaxAll\%), and the AUC is \tedAucMin--\tedAucMax{} at $k=1$, so additional members add little evidence and leave the criterion where it is. TED negatives are less similar in name than FOPPA's (median cosine \hardT{} against \hardF). Restricting both datasets to negatives with cosine at least 0.7 keeps the contrast. On FOPPA (\hsFn{} negatives) the false match rate rises from \hsFoneMin--\hsFoneMax\% to \hsFsixMin--\hsFsixMax\%, whereas on TED (\hsTn) it moves from \hsToneMin--\hsToneMax\% to \hsTsixMin--\hsTsixMax\% with no significant increase (Table~S13); neither country shows one separately (Table~S17). The FOPPA address ablation shows that the number of fields alone does not explain the difference. The snowball still replicates on TED in \tedSnSig{} of \tedSnN{} contrasts over models and sizes (Qwen2.5-32B: \tedSnThirtC\% to \tedSnThirtX\% at $k=4$). Accretion thus arises where single records are weak evidence of identity, the regime that motivates showing a profile at all.

\subsection{Setting the Operating Point}\label{sec:fix}
We sweep three remedies on the same 80 blocks (Fig.~S1). \emph{Size-calibrated thresholds} are fit per profile size on the 186 entities outside the blocks to keep the false match rate on hard negatives below $\tau\in\{0.05,\dots,0.3\}$. A \emph{global threshold} uses $t\in\{0.5,\dots,0.999\}$. A \emph{cluster audit} judges each member again against the founding member at sizes 4, 8 and 16 and evicts those it rejects.

Wherever the precision-recall curves overlap, the whole profile lies on or above the curve of the founding member for every model, so the two differ mainly in their operating point. In five-fold cross-validation over blocks (20 repetitions) we select, on the training blocks, the profile operating point with the highest recall whose precision is at least that of the founding member rule (a \emph{precision-matched threshold}). On the held-out blocks its precision differs from the founding member rule's by \fxDPmin{} to \fxDPmax{} points, with no 95\% interval excluding zero; this bounds any precision loss at \fxNiMin--\fxNiMax{} points (one-sided 95\% bounds). Its recall is \fxRRmin--\fxRRmax{} times higher for \fxRsig{} models (for Mistral-24B from \fxMisRone\% to \fxMisRtwo\%); Qwen3-32B keeps its recall (\fxQthreeRatio$\times$), since its founding member rule already has the highest recall. In clusters of eight or more members, \satOnThirt\% of Qwen2.5-32B's decisions (\satOnFourt\% for Qwen2.5-14B) have $P(\text{YES})\geq0.9999$, including \satOnThirtWrong\% of its wrong ones, so on the $P(\text{YES})$ scale of the sweep thresholds cannot separate them. BCubed scores~\cite{amigo2009comparison} agree (Table~S14). The blocks are confusable by construction and each fold calibrates on 64 labelled blocks of the same stream, so this is a stress test.

One global threshold is enough. In none of the \fxCalN{} comparisons at matched recall in which both curves are defined do thresholds per profile size change precision (paired block bootstrap). The audit raises the cost from \fxCallsProf{} to \fxCallsAud{} LLM calls per record and leaves $F_1$ unchanged (\fxAudProf{} on average, Table~S11); a stricter eviction rule moves precision by at most \fxSoftDP{} points. The effective remedy is therefore a precision-matched threshold set on labelled data and applied at admission. It also keeps the cost low: the whole profile needs \callPrMin--\callPrMax{} LLM calls per record, like the founding member, whereas pairwise max needs \callPmMin--\callPmMax{} (Table~S16).
\section{Discussion}

The accumulated profile acts as an operating-point variable. Every model in our study says YES more often as the profile grows, and for all models except Qwen3-32B the extra matches come at nearly the exchange rate that a lower threshold would give with a single member. A threshold of 0.5 therefore corresponds to a different operating point for every cluster size; in online linking, a precision-matched threshold restores the founding member's precision with more recall for five models (Section~\ref{sec:fix}). Where probabilities round to 1, as for Qwen2.5-32B at large profiles, thresholds have to be set on log-odds, and even there the large profile buys no recall at the false match rate of a single member. Comparing with individual records under an explicit aggregation rule keeps each decision tied to a record it can be checked against.

The snowball means that false merges concentrate on the intruder's entity. Contamination is best prevented at admission, by holding the operating point at the required precision; in our blocks this achieved what cluster audits afterwards did not (Section~\ref{sec:fix}).

The scope results locate the effect. Accretion appears when single records are weak evidence of identity, as with FOPPA names and locations even after addresses are added. On TED, where one record with its full address already identifies the company, recall with a single member is near its maximum. Further members do not move the criterion, even for similarly named negatives. It appears in pairwise, batched and candidate-selection prompts, whereas asking the model to partition a record set produces the opposite error. The error model therefore depends on the prompt format.

The result also changes what ER theory can assume about LLM matchers. Guarantees for oracle-based ER assume an error model that depends on the record pair~\cite{firmani2016online,galhotra2018robust}; for an LLM shown a profile they need one indexed by cluster size. Our design measures that index directly and needs only a registry identifier as ground truth; with a fixed query, a nested reference set and external ground truth, it applies equally to other LLM judgments made against accumulated references. FOPPA effect sizes refer to entities with at least 17 agents, the regime in which profiles reach 16 members. The evidence spans company records from three countries, eight models from four developers and four prompting formats.

\section{Conclusion}

When an LLM matcher is shown a cluster profile, its decision criterion drifts toward acceptance as the cluster grows. On FOPPA suppliers with SIREN ground truth, sixteen verified members raised false matches on similarly named companies by \dfmrMin{} to \dfmrMax{} percentage points in six open models, with little gain in discrimination. In two Claude models scored by sampled answers the rise was \clSonD{} and \clHaiD{} points. One false merge made further false merges from the same legal entity more likely. The shift holds across fields, prompt wordings and two published matching formats, and it vanishes where a single record already identifies the company. For company records like these, incremental LLM-based ER should set the profile matcher's threshold on labelled data for the required precision. Code, seeds, model inputs, outputs and the online supplement (Tables S1--S18, Fig.~S1) are available at \url{https://github.com/xuemian168/profile-accretion} (tag \texttt{mita2027-v1}).

\ifanon\else
\section*{Acknowledgment}
We thank Yifei Zhou and Yuxuan Cai (Compliance0 Pty Ltd) for discussions on procurement supplier data.
\fi

\bibliographystyle{IEEEtran}
\bibliography{refs}

\end{document}